# TOWARDS EFFICIENT NEURAL NETWORKS ON-A-CHIP: JOINT HARDWARE-ALGORITHM APPROACHES

*Xiaocong Du[1], Gokul Krishnan[1], Abinash Mohanty[1], Zheng Li[2], Gouranga Charan[1], Yu Cao[1]\**
[1]School of ECEE, [2]School of CIDSE, Arizona State University, Tempe, AZ 85287, USA
\*Corresponding Author's Email: ycao@asu.edu

## ABSTRACT

Machine learning algorithms have made significant advances in many applications. However, their hardware implementation on the state-of-the-art platforms still faces several challenges and are limited by various factors, such as memory volume, memory bandwidth and interconnection overhead. The adoption of the crossbar architecture with emerging memory technology partially solves the problem but induces process variation and other concerns. In this paper, we will present novel solutions to two fundamental issues in crossbar implementation of Artificial Intelligence (AI) algorithms: device variation and insufficient interconnections. These solutions are inspired by the statistical properties of algorithms themselves, especially the redundancy in neural network nodes and connections. By Random Sparse Adaptation and pruning the connections following the Small-World model, we demonstrate robust and efficient performance on representative datasets such as MNIST and CIFAR-10. Moreover, we present Continuous Growth and Pruning algorithm for future learning and adaptation on hardware.

## INTRODUCTION

Rapid development of convolutional neural networks (CNNs) have enabled multiple applications such as image classification [1], object detection [2] and speech recognition [3]. However, these tasks currently depend on highly over-parameterized networks. With excessive cost in memory and computational resources, it is difficult to deploy them on embedded hardware. Implementation of a neural network on an integrated circuit, i.e., Network-on-Chip (NoC), as shown in Figure 1, usually involve many crossbar arrays to execute the computations. Though promising, NoC is facing challenges including but not limited to device variations and insufficient connections.

Current solutions to tackle device variation include "Closed-Loop-on-Device" (CLD) [4] and "Open-Loop-off-Device" (OLD) [5]. CLD repeatedly programs and senses on-chip to complete the function of gradient descent. However, it is time-consuming due to the multiple times of writing. OLD executes the flow that pre-trained networks are used to calculate the resistance of the devices, and then programs and senses by Read-Verify-Write (R-V-W), till the resistance converges to the desired values. The limitation of OLD is that the CNNs must be trained from scratch for each chip. Moreover, inference accuracy in all the above methods is limited by the quantization error (number of levels) of the RRAM device.

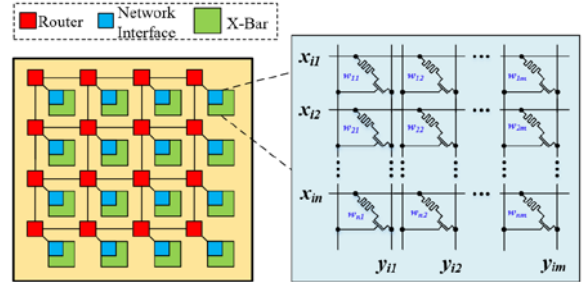

*Figure 1: Structure of Network-on-Chip (NoC) and crossbar array*

Network pruning, a prevalently-recognized solution to simultaneously accelerate and compress CNN models, removes less significant weights from a pre-trained and over-parameterized network. Network pruning usually follows a three-step scheme which trains an extensive network from scratch, removes unimportant connections or neurons according to a predefined ranking score and fine-tunes the rest. However, training a large network from scratch and pruning during training could be sub-optimal as it introduces redundancy.

Recognizing the expensive cost and non-ideal performance of the methods above, this work proposes solutions from the perspective of algorithm-hardware co-design:

1. A fundamentally new approach, Random Sparse Adaptation (RSA), which mitigates the impact of device variance with high effectiveness and efficiency.
2. A novel training scheme inspired by Small-World models, that tackles insufficient connections and improve efficiency of on-chip training and inference.
3. A novel algorithm, Continuous Growth and Pruning (CGaP), that further achieves adaption of network structure size and inspires future learning.

## DEVICE VARIATION AND SOLUTIONS
### Non-ideal effects in a RRAM device

A real-world RRAM device only has fixed levels, limited by on- and off- resistance, variations, etc. Write variation follows a lognormal distribution [4], where stuck-at-faults occur when a device is always at either high resistance state (SF1) or low resistance state (SF0). In an array, SF1 and SF0 are assumed to affect 9.04% and 1.75% of the devices, respectively [5]. Due to these non-idealities, the distribution of pre-trained model parameters is distorted when they are programmed to an RRAM array (shown in Figure 2), resulting in significant degradation of inference accuracy, especially for complex datasets such as CIFAR-10.

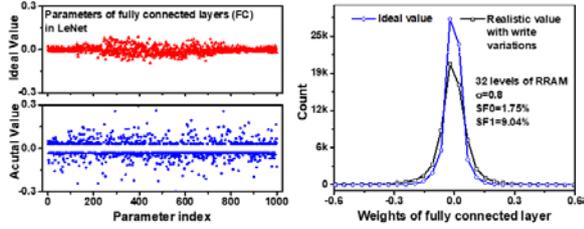

*Figure 2: The deviation of model parameters after Write to the realistic RRAM array.*

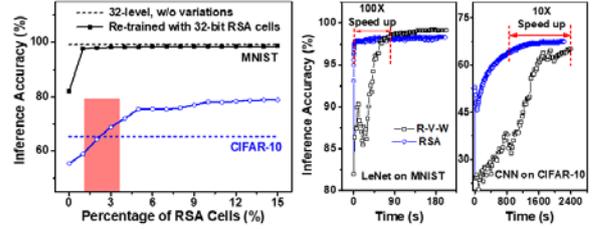

*Figure 4: Left: A small amount of 32-bit RSA cells effectively improves the accuracy; Right: RSA rapidly recovers the accuracy, achieving 10-100x speedup over R-V-W.*

**Solution: Random Sparse Adaptation (RSA)**

To recover the accuracy loss caused by device variance, R-V-W is commonly used. However, R-V-W is time-consuming: executing gradient descent on-chip while writing the RRAM cells is slow and inefficient. The RSA method proposes to randomly select a small subset of cells and duplicate them in a separate on-chip memory (Figure 3 left), effectively enhancing the programming speed in adaptation. The random selection is further regularized in the sense that an equal number of cells will be selected from each row and column of the original array, such that the selected cells can be compiled into a rectangular array for a compact footprint of the RSA array and the periphery circuitry. The cell positions are still random, and the random connections between RSA and RRAM input/output are hard-wired.

Figure 3 (right) presents the architecture, design and operation of the proposed RSA scheme. First, pre-trained models are programmed to RRAM array and the on-chip RSA cells are initialized from random-normal distribution.

During the feedforward inference, the input to the layers is passed to both RRAM array and parallel on-chip RSA memory. The output from both are added to generate the overall layer output. During back-propagation for training, RRAM is only Read and the output is combined with RSAs to calculate the gradient, which is then used to adapt the RSA cells. As the parameters on RRAM array are masked as non-trainable in this method, no write operation is necessary for RRAM cells. Only the on-chip RSA cells are updated to tune the overall network in the direction to improve the accuracy.

**Performance of RSA**

To demonstrate the efficacy of RSA, two representative datasets are used, MNIST [6] for handwritten digit recognition and CIFAR-10 [7] for more complicated image recognition.

Figure 4 (left) shows that using RSA, only a tiny portion of parameters (<5%) is required to compensate the effect of device variation and stuck-at-faults on inference accuracy. With gradient descent operating over high-precision RF cells, much better accuracy can be achieved than that with 32-level RRAM only. For instance, in the 9-layer CNN for CIFAR-10, the addition of 5% RSA cells boosts the accuracy by more than 10%, which is very significant for this task. Figure 4 (right) compares the improvement in accuracy and the time needed, between RSA and R-V-W. Leveraging the robustness of the algorithm, rather than device-level precision, RSA achieves higher accuracies at a much faster speed. The speed-up is in the range of 10-100X, depending on the R/W time of RRAM and the number of convolutions in the algorithm.

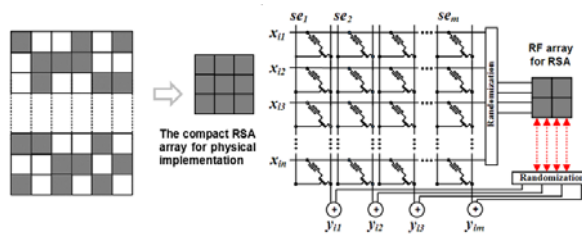

*Figure 3: The network structure, design and flow using RF for RSA cells. The backpropagation only goes through RF cells. RRAM is Read only.*

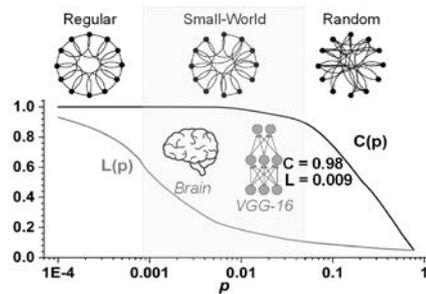

*Figure 5: The classical LC curve to illustrate the structural property. The Small-World region exhibits high Clustering Coefficient C and low Characteristic Path Length L.*

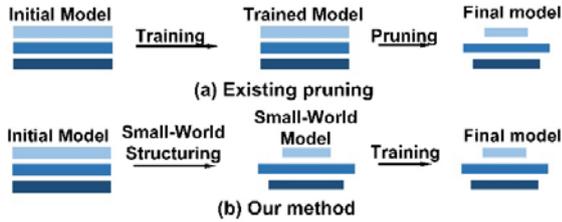

*Figure 6: (a) Regular pruning method (b) Small-World method. The Small-World method starts from a sparse network for training and achieves a sparse and accurate model comparable or better than existing methods.*

## INTERCONNECTION PROBLEM
### Insufficient on-chip connections

DNNs are excessively over-parameterized and redundant, resulting in a larger memory and interconnection overhead. Current pruning mechanisms exploit the over parametrization by removing redundant connections in the network [8]–[10]. But the existing pruning methods do not incorporate the intrinsic structural property of the network, thus lack of the control of the location to be pruned.

### Solution: Small-World based pruning mechanism

Small-World structure is a semi-random structure that can be characterized by two properties, namely, high clustering coefficient C and low characteristic path length L. The classical Small-World definition [11] introduces L and C through the example of friendship networks, where L is the average number of friendships in the shortest chain connecting two people and C reflects the extent to which friends of one are also friends of each other. We take a cue from this to define L and C for Small-World Neural Networks as:

(1) Friendship is defined as a weighted connection between two neurons/feature maps in a class and each class as a friendship network.

(2) For fully connected layers Characteristic Path Length L is the number of neurons connected to the class (friendships per class), averaged over all the classes. Clustering Coefficient C is the number of classes each neuron is contributing to (friends who are also friends of each other), averaged over all neurons.

(3) For convolutional layers, Characteristic Path Length L is the number output features contributing to the class (friendships per class), averaged over all the classes. Clustering Coefficient C is the number of classes each output feature contributes to, averaged over all output features.

The Small-World based pruning incorporates the structural property (L and C, shown in Figure 5) of the network into the pruning mechanism. Figure 6b shows the flow of the proposed pruning mechanism. Figure 7 gives the L and C plots for the Small-World model of the VGG-16 network [12]. For increased sparsity threshold Θ at a constant random probability p, the L reduces considerably while a high C is maintained, thus following the Small-World phenomenon for the network. The Small-World pruning provides the following advantages: (1) A pre-defined structure which gives prior knowledge of the nodes and connections in the network; (2) Clustering of features resulting in a locally dense and globally sparse model that enables efficient implementation on ReRAM based hardware platforms [13]; (3) A scalable sparse model with lower memory and interconnection cost.

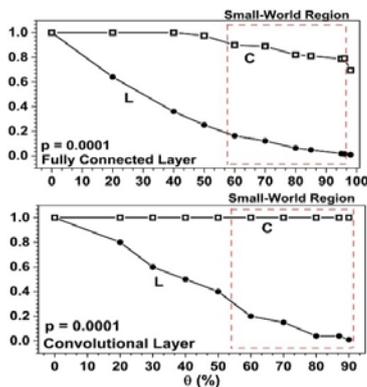

*Figure 7: Variation in L and C for the fully connected and convolutional layer of a Small-World VGG-16 on CIFAR-10 with increase in sparsity θ of the layer. The curve shows high clustering coefficient and low Characteristic Path Length at high θ values for p = 0.0001.*

Table. 2 summarizes the overall result of the Small-World based pruning approach. It achieves a model compression of 97.7% for LeNet-5 [6] on MNIST, higher than existing methods. The number of parameters achieved by the Small-world model of LeNet- 5 is lesser than that of [14] for similar accuracy. For Modified VGG-16 [15] on CIFAR-10 we obtain a lower number of parameters for a near baseline accuracy as compared to [16]. For regular VGG-16 on CIFAR-10 we achieve a 90.8% reduction in the number of parameters for 93.24% accuracy.

TABLE I. OVERALL RESULTS

| Model | Accuracy | Param. | Pruned |
|---|---|---|---|
| LeNet-5 | | | |
| Baseline | 99.29 % | 431K | - |
| Pruning [12] | 99.20 % | 10.7K | 97.51 % |
| Small-World | **99.18 %** | **9.9 K** | **97.70 %** |
| VGG-16 | | | |
| Baseline | 92.93 % | 15.3M | - |
| Pruning [14] | 93.40 % | 5.4M | 64.00 % |
| Small-World | **93.10 %** | **5.3M** | **66.00 %** |

## FUTURE: CONTINUAL GROWTH
**Importance of Learning Network Structure**

A canonical pipeline to develop accurate and efficient Deep neural networks (DNNs) is: (1) pre-define the structure of a DNN with redundant learning units (such as filters and neurons) with the goal of high accuracy; (2) prune redundant learning units with the purpose of efficient inference. We argue it is a paradox to increase redundancy in training but to decrease redundancy for efficient inference.

The paradox can result in over-fitting during training and an excessive computation cost. The fixed structure (i.e., width and depth of the layers) further results in poor adaption to dynamic tasks, such as incremental and lifelong learning. In contrast, structural plasticity plays an indispensable role in mammalian brains to achieve compact and accurate learning. As shown in the previous section, existing pruning methods have certain drawbacks, in addition to which there are further structural implications like, (1) in the process of training, pruning only discards less important weights at the end of training but never strengthens important weights and nodes; (2) the over-parameterization could be caused by the incapability of a fixed structure to predict the network size at the beginning of training. An over-sized network is usually selected to guarantee training accuracy. Therefore, we present a novel algorithm, Continuous Growth and Pruning (CGaP), to address such limitations without performance degradation.

**Solution: Continuous Growth and Pruning (CGaP)**

Instead of training an over-parameterized network from scratch, CGaP (as shown in Figure 8) starts the training from a small network seed, size of which can be as low as 0.1 % of a baseline model. In each iteration of growth, CGaP sorts neurons and filters locally by their saliency in the reduction of the loss function. Based on the saliency score, important learning units are favored, and new-born units are added, leading to a more explicit learning purpose as well as an increased model capacity. A filter and neuron-wise pruning is then executed on the post-growth model (with peak capacity) based on pruning metrics, which offers a significantly sparse inference model with a structural growth and learn implication.

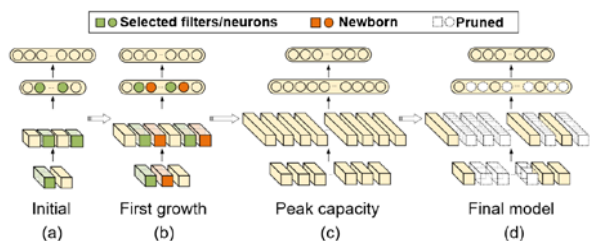

*Figure 8: Algorithm of proposed CGaP flow. CGaP (a) starts the learning from a small network seed, instead of an over-parameterized structure, (b) continuously adds important learning units and widens each layer, (c) obtains a peak network, (d) prunes to reach a compact model.*

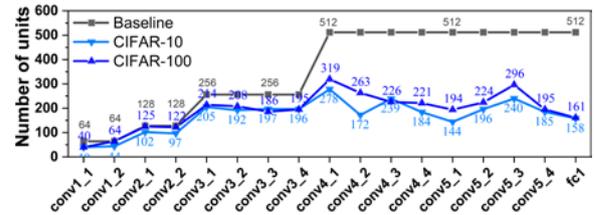

*Figure 9: CGaP can learn the necessary width for each layer given different datasets.*

Experiments on CIFAR-10 and CIFAR-100 [7] prove that CGaP can learn the network structure and size for datasets in different complexity. The network to learn CIFAR-10 is not necessarily as large as the one for CIFAR-100, as shown in Figure 9.

## ACKNOWLEDGEMENTS

This work was supported in part by the Semiconductor Research Corporation (SRC) and DARPA.